\documentclass[journal]{IEEEtran}
\usepackage{multirow}
\usepackage{amssymb}
\usepackage{graphicx}
\usepackage{amsmath}
\ifCLASSINFOpdf
\else
\fi
\begin{document}
%
\title{Spatio-temporal Gait Feature with Global Distance Alignment}

%
%
%

%
%

\author{Yifan~Chen, Yang~Zhao, Xuelong~Li,~\IEEEmembership{Fellow,~IEEE}
\thanks{The authors are with the School of Artificial Intelligence, OPtics and ElectroNics (iOPEN), Northwestern Polytechnical University, Xi'an 710072, P. R. China, and with the Key Laboratory of Intelligent Interaction and Applications (Northwestern Polytechnical University), Ministry of Industry and Information Technology, Xi'an 710072, P. R. China.}
\thanks{Corresponding author: Xuelong Li.} 
}


%



\maketitle

\begin{abstract}
  Gait recognition is an important recognition technology,                                        
   because gait is not easy to camouflage and does not need cooperation to recognize subjects. 
   However, many existing methods are inadequate in preserving both temporal information and
   fine-grained information, thus reducing its discrimination.
    This problem is more serious when the subjects with similar walking postures are identified.
     In this paper, we try to enhance the discrimination of spatio-temporal gait features from two aspects:
     effective extraction of spatio-temporal gait features and reasonable refinement of extracted features.
   Thus our method is proposed, it consists of Spatio-temporal Feature Extraction (SFE)
    and Global Distance Alignment (GDA).
    SFE uses Temporal Feature Fusion (TFF) and Fine-grained Feature Extraction (FFE) to
    effectively extract the spatio-temporal
    features from raw silhouettes. GDA 
    uses a large number of unlabeled
    gait data in real life as a benchmark
    to refine the extracted spatio-temporal features.  GDA can make the extracted features
     have low inter-class similarity and high intra-class similarity,
     thus enhancing their discrimination.  
    Extensive experiments on mini-OUMVLP and CASIA-B have proved that
    we have a better result than some state-of-the-art methods. 
\end{abstract}


%
\IEEEpeerreviewmaketitle

\section{Introduction}
%
%
%
%
\IEEEPARstart{D}{ifferent} from iris, fingerprint, face, and other biometric features, gait
 does not need the cooperation of subject in the process
 of  recognition, and it can recognize the subject at a long distance in uncontrolled scenarios.
  Therefore,
 gait has broad applications in forensic identification, video surveillance, 
 crime investigation, etc \cite{marin2021ugaitnet,choi2019skeleton}. As a visual identification task, its goal is to learn
 distinguishing features for different subjects. 
  However, when learn spatio-temporal features from raw gait sequences, gait recognition is often disturbed by
  many external factors, 
  such as various camera angles, different
   clothes/carrying conditions \cite{he2018multi,li2017a,li2018person}.

 Many deep learning-based methods 
 have been 
 proposed to overcome these problems \cite{su2020deep,liao2020attention,ma2020deep,zhang2019gait,Zhang_2019_CVPR,castro2020multimodal}.
  DVGan uses GAN to produce the whole view space, which views angle from 0° to 180° with 1°
  interval
 to adapt the various camera angles \cite{liao2020dense}.  GaitNet uses Auto-Encoder as their framework to
  learn the gait-related information
 from raw RGB images.
 It also uses LSTMs to learn
 the changes of temporal information to overcome the different clothes/carrying conditions \cite{song2019gaitnet}.
 GaitNDM encodes body shape and boundary curvature into a new feature descriptor to increase
  the robustness of gait representation \cite{el2017gait}.
 GaitSet learns identity information from the set, which consists of independent frames, 
 to adapt the various viewing angles and different clothes/carrying conditions \cite{chao2019gaitset}. 
 Gaitpart employs partial features for human body description to enhance fine-grained learning \cite{Fan_2020_CVPR}. 
 One robust gait recognition method combines multiple types of sensors to enhance the richness of gait information \cite{zou2017robust}.  
 Gii proposes a discriminant projection method based on list constraints to solve the perspective variance problem in cross-view gait recognition \cite{zhang2017gii}.

 Previous methods have performed well in extracting temporal or spatial information, but still fall short in
 extracting spatio-temporal features at the same time.
 This problem is even worse when identifying subjects with similar walking postures.
  To solve this problem, we came up with a novel
  method: Spatio-temporal Gait Feature with Global Distance Alignment,
   which consists of Spatio-temporal Feature Extraction (SFE) and Global
  Distance Alignment (GDA). SFE includes Temporal Feature Fusion (TFF) and Fine-grained Feature Extraction(FFE). 
   The proposed method enhances the discrimination of spatio-temporal gait features from two aspects:
  effective extraction of spatio-temporal gait features and reasonable refinement of extracted features.
  For effective extraction of spatio-temporal gait features, we first
  use TFF to fuse the most representative temporal features
  and then use FFE to extract fine-grained features from the most
   representative temporal features.
  After the above operations, spatio-temporal features of raw gait silhouettes can be fully extracted. 
  For reasonable refinement of extracted features, we try to use the feature distribution of real-life 
  unlabeled gait data as a benchmark to refine the extracted spatio-temporal gait features,
  this operation can further increase the
  intra-class  similarity and inter-class dissimilarity of the extracted spatio-temporal gait features.

We will describe the advantages of the proposed method from the following three aspects:

\begin{itemize}
  \item We propose a concise and effective framework named Spatio-temporal Feature Extraction (SFE), which firstly uses
  TFF to fuse
   the most representative temporal features, and then uses FFE to extract fine-grained features from the most
   representative temporal features. SFE is also a lightweight network model that  has fewer network parameters than GaitSet, GaitPart and other 
   state-of-the-art methods.
  \item We propose a Global Distance Alignment (GDA) technique, which uses the feature distribution of real-life
   unlabeled gait data as a benchmark to refine the extracted spatio-temporal gait features. GDA can greatly
  increase the intra-class similarity and inter-class dissimilarity of extracted spatio-temopral gait features,
  thus enhancing their discrimination.
  \item Extensive experiments on CASIA-B and mini-OUMVLP have proved 
  that our method have better performance than other state-of-the-art methods. It is worth noting that
  the proposed method
  achieves an average rank-1 accuracy of 97.0$\%$ on the CASIA-B gait dataset
  under NM conditions.
  \end{itemize}

  \begin{figure*}
    \centering
    \includegraphics[width=17cm]{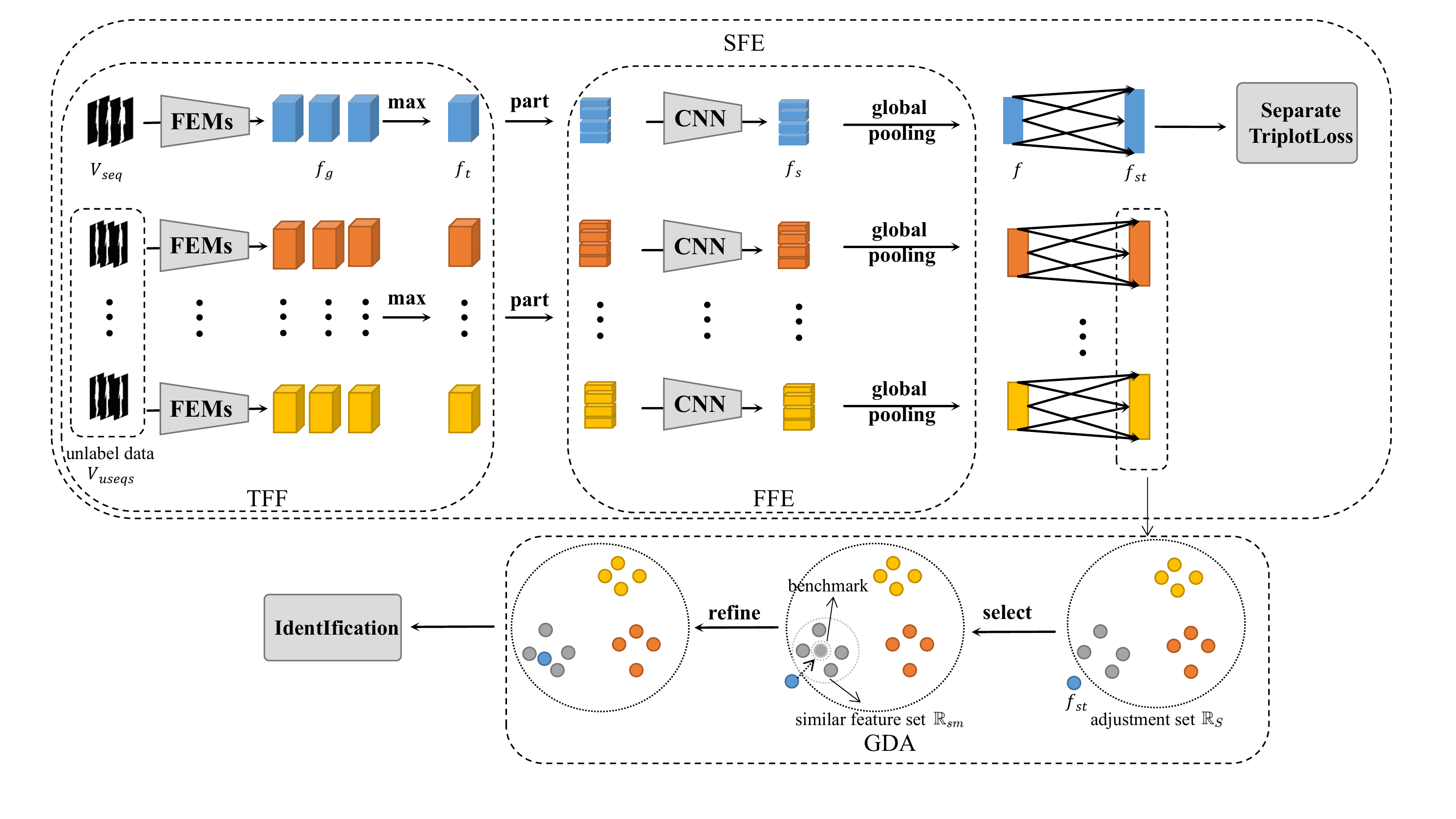}
    \caption{The framework of our method.  It is clearly observed that our framework consists of two parts:
     Spatio-temporal Feature Extraction (SFE) module and Global Distance Alignment (GDA). The  SFE  module  is
   mainly  the  optimization  of  neural  network  structure. GDA is the refinement for the extracted
   spatio-temporal gait features. At the bottom of the figure, in order to observe the distribution
    of features, we use circles to represent the extracted spatio-temporal gait features.}
    \label{fig}
  \end{figure*}

\section{Related Work}
 This section focuses on the work related to gait recognition. It first introduces the differences between
 appearance-based and model-based methods, and then mainly introduces the temporal feature extraction model
 and spatial feature extraction model in appearance-based methods.

{\bf Gait recognition.} Current gait recognition methods can be categorized as 
 model-based methods \cite{an2020performance,liao2020model,yan2018spatial,kastaniotis2016pose,chen2018image}
 and appearance-based methods \cite{yu2019weizhi,wang2019ev,rida2019towards,ma2017general,gadaleta2018idnet}. 
 Appearance-based methods use CNNs directly to learn the spatio-temporal features of the original gait sequences
 and then use feature matching to identify the gait sequence \cite{ben2019general}.
 Model-based methods use a new representation to replace the original gait silhouettes as input.
 A representative model-based method is JointsGait \cite{li2020jointsgait},
 which uses the human joints of raw gait silhouettes to create gait graph structure
  and then extracts the spatio-temporal
 features from the gait graph structure by Graph Convolutional Network (GCN). However,
 when using this method to express the silhouettes in gait sequence,
  it often loses a lot of important detailed information and 
 increases the difficulty of recognition. Other
 model-based methods also encounter this problem,
  so appearance-based methods have become the most mainstream gait recognition methods at present
 \cite{liao2020attention,sepas-moghaddam2020gait,lin2020learning,ma2020deep,zhao2020learning,ji2020sgap,fu2019horizontal}. In this paper, the methods 
  we mentioned later both belong to the appearance-based methods.

   The efficiency of spatio-temporal feature extraction 
   is an important factor to measure the quality of appearance-based methods\cite{deng2017fusion},
    it will greatly affect the accuracy of recognition.
   Spatio-temporal feature extraction module can be divided into two parts:
   temporal feature extraction module\cite{wang2010chrono} and spatial feature extraction module. 
   {\bf{1)}} For the temporal feature extraction module,
  there have deep learning-based methods and traditional methods.
  The traditional methods first compress the original gait silhouettes into a single image and then use a convolutional
   neural network to extract the spatio-temporal features of that image \cite{han2005individual,shiraga2016geinet}.
 Although these traditional methods are simple,
   the following researchers have found that they do not preserve temporal information well and
  try to use deep 
  learning-based methods to extract temporal features. 
  LSTM \cite{zhang2019classification,feng2016learning,sepas2020facial} uses a repeated neural network module to preserve and extract the temporal information from raw
   gait sequences.
   GaitSet\cite{chao2019gaitset} observes that even if the gait sequences are shuffled,
  rearranging them into correct order is not difficult, and then uses shuffled silhouettes to
   learn spatio-temporal information to ensure 
  the adaptability of various gait sequences.
  But both LSTM and GaitSet have some disadvantages: they have complex network
   structures and calculation processes. In this paper, we try to use several parallel convolution layers
    to extract global features from 
   raw gait silhouettes and use a simple max operation to fuse the most representative temporal feature.
    This can simplify
   the network structure and use less calculation to effectively extract the temporal feature. It is worth
    noting that our method only uses 4 convolution layers and 2 pooling layers in total.
   {\bf {2)}} For the spatial feature extraction module, in \cite{li2008gait}, the idea of partial is introduced into gait
    recognition, and it is considered that different human parts will play different roles in the
     recognition of identity information. Therefore, the human body is divided into seven different components,
      and the influence of different parts on gait recognition is explored
    by removing the seven components in the average gait image and observing the change of recognition rate.
    It lays a foundation for the current use of the partial idea.
   GaitPart uses this idea in the field of deep learning to  describe the human body 
  to fully extract the fine-grained information, it divides multi-layer features into
   blocks to extract fine-grained features
  and achieves good results. 
  However,
  GaitPart uses the idea of partial in shallow features and divides
   multi-layer features into blocks, which increases the
  complexity of the network. In this paper, we divide the high-level features only once to extract the fine-grained
  feature, and only a single convolution layer can get good performance. This interesting idea also greatly
   simplifies the network structure.

   We fully consider the shortcomings of previous gait recognition methods in spatio-temporal feature extraction
   and propose our method:
   Spatio-temporal Gait Feature 
   with Global Distance Alignment. It first designs a concise and effective spatio-temporal gait feature extraction (SFE) model
   to ensure the effective extraction of spatio-temporal gait features, and then uses the global distance
   alignment (GDA) technique to increase the intra-class similarity and inter-class dissimilarity of the extracted
   spatio-temporal gait features. Our method can effectively enhance the discrimination of gait sequence to ensure
   the accuracy of gait recognition.

\section{Our method}

 We will introduce our method from three aspects in this section,  
 which includes the overall pipeline, Spatio-temporal Feature Extraction (SFE)
 module and Global Distance Alignment (GDA).
 SFE mainly includes
 two parts:
 Fine-grained Feature Extraction (FFE) and
 Temporal
 Feature Fusion (TFF). The framework of the proposed
  method is presented in Fig. \ref{fig}.

\subsection{Pipeline}
 As presented in Fig. \ref{fig}, our network mainly consists of two parts: SFE module
 and GDA module. SFE module is mainly the optimization 
 of neural network structure, it can effectively extract the spatio-temporal gait features of raw 
 gait sequences by a flexible network structure.
  GDA is a post-processing method, which uses a large number of unlabeled gait data in real life as a benchmark
 to refine the extracted spatio-temporal gait features.
  GDA can further enhance the discrimination of extracted gait features.
 It is worth noting that GDA is used in test phase, and it uses the trained SFE model as an extractor to
 extract the adjustment set.
    
     At the part of SFE, we first input some raw gait silhouettes
     to TFF module frame by frame.  TFF can learn the most representative temporal gait feature $f_t$ from the
    original gait sequence $V_{seq}$. 
    \begin{equation}
      f_t = TFF(V_{seq}).
      \end{equation}
   The most representative temporal gait feature $f_t$ are then sent to FFE module to learn their
    fine-grained feature $f_s$. 
    \begin{equation}
      f_s = FFE(f_t).
      \end{equation}
    After fine-grained features $f_s$ are extracted, a global pooling layer is used to remove redundant information
    of feature $f_s$ to obtain feature $f$. 
    Then the fully connected layer is used to integrate the spatio-temporal information
    of feature $f$ into $f_{st}$. Finally, the Separate TripletLoss is used as a constraint to optimize network
    parameters. 

    At the part of GDA, we first use the trained SFE model to extract the spatio-temporal gait features of
    unlabel data $V_{seqs}$ as the adjustment set
    $\mathbb{R}_S$.
    \begin{equation}
      \mathbb{R}_S = SFE(V_{useqs}).
      \end{equation}
      Then we select some similar spatio-temporal gait features from $\mathbb{R}_S$
      as similar feature set $\mathbb{R}_{sm}$ to refine the
      extracted feature $f_{st}$
      to make it have high
      intra-class similarity and low inter-class similarity, thus enhancing its discrimination.

\subsection{Spatio-temporal Feature Extraction model}
 SFE model is mainly the optimization of neural network.
 It ensures that the spatio-temporal gait features are fully extracted while using as few network parameters
 as possible. We will introduce it from Temporal Feature Fusion (TFF) and Fine-grained Feature Extraction (FFE).

\subsubsection{Temporal Feature Fusion (TFF)}

 TFF 
 consists of several parallel FEMs and a max pooling layer, where the parameters of FEMs
 are shared. FEMs are used to extract the global
 features $f_g$ from original gait sequence $V_{seq}$,
 they can fully extract the important information of each frame.
 Max pooling is used to select the most representative temporal feature $f_t$ from the global features $f_g$.
 We will introduce the structure of FEM and why do we use Maxpooling to fuse the most representative temporal
 feature $f_t$ next.

\begin{table}[h]
  \centering
  \renewcommand\arraystretch{1.3}
  \setlength{\tabcolsep}{1.6mm}
  \caption{The structure of Feature Extraction Module. In, Out, Kernel and 
  Pad represent the channels of input and output, kernel size and padding of
  Conv2d.}
  \label{tab1}
  \begin{tabular}{|c|c|c|c|c|c|}
  \hline
  \multicolumn{6}{|c|}{Feature Extraction Module}                                 \\ \hline      
  Block                   & Layer      & In      & Out     & Kernel     & Pad     \\ \hline
  \multirow{3}{*}{Block1} & Conv2d (C1)    & 1       & 32      & 5          & 2       \\ \cline{2-6} 
                          & Conv2d (C2)     & 32      & 32      & 3          & 1       \\ \cline{2-6} 
                          & \multicolumn{5}{c|}{Maxpool, kernel size=2, stride=2} \\ \hline
  \multirow{3}{*}{Block2} & Conv2d (C3)     & 32      & 64      & 3          & 1       \\ \cline{2-6} 
                          & Conv2d (C4)     & 64      & 128      & 3          & 1       \\ \cline{2-6} 
                          & \multicolumn{5}{c|}{Maxpool, kernel size=2, stride=2} \\ \hline

  \end{tabular}
  \end{table}

  As presented in Tab. \ref{tab1}, FEM uses 4 Convolutional layers to extract the global features of raw 
  silhouettes and 2 Max pooling
  layers to select the important information of these global features. These operations can fully extract the
  global features $f_g$ from original gait sequence.
  \begin{equation}
    f_g = FEM(V_{seq}).
    \end{equation}
  We also find that only some features we extracted
   play an outstanding role in describing
  human identity.  We try to use various operations: max, mean, medium 
   to fuse the representative frame from extracted global features 
   and then we find that using max operation can better preserve human identity information.
    The max operation is defined as:
  \begin{equation}
    f_t = Maxpooling(f_g),
    \end{equation}
    where $f_g$ are the global features extracted from raw gait sequences by several parallel FEMs,
     $f_t$ is the most representative
    temporal feature fused from  $f_g$ by max operation.

\subsubsection{Fine-grained Feature Extraction (FFE)}

 In the former literature, some researchers have found that using partial features to
 describe the human body has an outstanding performance. However, previous researches both analyze shallow partial
 features to describe the human body \cite{verlekar2018gait} and it needs
  to be partitioned many times in different layer features 
 to extract fine-grained features, which will make the structure of the network more complex.
 So we try to extract the fine-grained features from deep layers to simplify network structure.
 It can ensure that we can fully extract fine-grained features while minimizing network parameters.
 We sequentially divide the extracted most representative temporal feature $f_t$ into 1, 2, 4, and 8 blocks to
  learn the fine-grained information,
  which the most representative temporal feature $f_t$ is a high-level feature.
  We can observe that when we use 4 blocks to learn the fine-grained features $f_s$, it has the best performance. 
  The results show that when we learn fine-grained information, excessive blocking can not make the fine-grained
   information learn more fully, but may ignore the relationship between adjacent parts.
  We also try to use deeper convolutional layers to learn fine-grained information again.  But in most cases,
   its performance is
  not as good as the 
  single Convolutional layer. The results strongly prove that when
   using the partial idea in high-level features, it does not need too deep network structure to
    learn fine-grained information. 

 Specifically, we first divide the most representative temporal feature $f_t$
 into 4 blocks. Then use a single Convolutional
 layer to learn the fine-grained feature $f_s$. After this,
 Global Pooling is used to reduce the redundant information of $f_s$.
 The formula of Global Pooling is shown as:  
\begin{equation}
  f = Avgpooling(f_s) + Maxpooling(f_s).
  \end{equation}
  These two pooling operations are both applied to the $w$ dimension. Through this operation,
   our feature changes from three-dimensional data containing $c$, $h$, and $w$ to two-dimensional data containing
    $c$ and $h$.  
    Finally, we use the Fully Connected Layer to integrate all the previously learned information to
    obtain the spatio-temporal gait feature $f_{st}$.

    Gait recognition is mainly based on feature matching to identify the identity of the subject.
    Therefore, the low similarity between
    different subjects and 
   high similarity between the same subject
   are important factors to determine the recognition accuracy.
   The hard triplet loss
   has a good performance on reducing inter-class similarity and increasing intra-class similarity,
    so we use it as a constraint to optimize network parameters.
    Hard triplet loss is formulated as: 
 \begin{equation}
  \begin{aligned}
  L_{trip} = m&ax( m + max(d_{i}^+) - min(d_{i}^-), 0),\\
  &d_{i}^+ = \mid\mid{f_{i}^{a_i}-f_{i}^{p_i}}\mid\mid_2^2,\\
  &d_{i}^- = \mid\mid{f_{i}^{a_i}-f_{i}^{n_i}} \mid\mid_2^2,
  \end{aligned}
  \end{equation}
 where $d_i^+$ is a measure that represents the dissimilarity of the positive sample and anchor ($a_i$ and $p_i$), 
$d_i^-$ is a measure that represents the dissimilarity of the negative sample and anchor ($a_i$ and $n_i$).
 And we use the Euclidean norm to get the values of $d_i^+$ and $d_i^-$.
 We use the min value of $d_i^-$ and the max value of $d_i^+$ as a proxy to calculate loss can be more 
 effective in reducing inter-class similarity and increasing intra-class similarity.

\subsection{Global Distance Alignment}

 Triplet loss can increase intra-class similarity and decrease inter-class similarity of subjects 
 in most cases. However, when meeting subjects with similar walking styles,
 triplet loss do not ensure the discrimination of their features. We try to use a
  post-processing method to refine the extracted spatio-temporal gait features
 to make them more discriminative. Global distance alignment technology is introduced to solve the problem.
 It pulls in the gait features between the same subjects to increase the intra-class similarity,
 and separates the gait features between different subjects to increase the inter-class dissimilarity. 

\begin{figure*}[htbp]
      \centering
      \includegraphics[width=17cm]{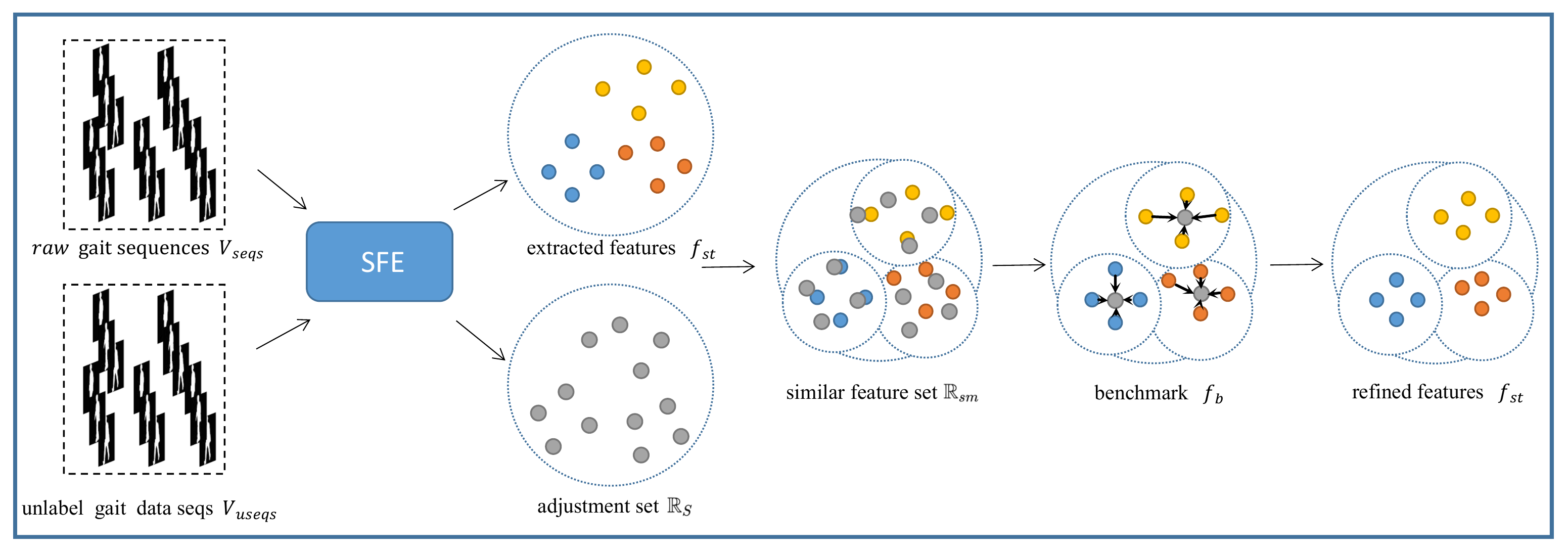}
      \caption{Raw Gait sequences $V_{seqs}$ and unlabel  gait data $V_{useqs}$  both input to trained SFE module,
       we can see original inter-class dissimilarity and the intra-class similarity of features $f_{st}$ 
       are small. The spatio-temporal features extracted from $V_{useqs}$ are used as adjustment set $\mathbb{R}_S$.
       When the adjustment set $\mathbb{R}_S$ is obtained, we will select the similar features as the similar
        feature set $\mathbb{R}_{sm}$, and finally select the appropriate benchmark $f_b$ from $\mathbb{R}_{sm}$ to
         refine the extracted features $f_{st}$.
        We can see after this operation, the inter-class dissimilarity and intra-class similarity of feature
        $f_{st}$ are increasing.}
      \label{fig1}
  \end{figure*}

 First, we will find a large number of  unlabeled gait data $V_{useqs}$ in  real  life. 
  Then, we use the trained SFE module to extract spatio-temporal features
 of these unlabeled gait data. These extracted features are used as an adjustment set $\mathbb{R}_S$. 
  And then, we calculate the distance between the features in $\mathbb{R}_S$ and the extracted
   spatio-temporal features $f_{st}$, and select  some  features $f_{sm}$ similar  to  $f_{st}$  from  
  $\mathbb{R}_S$ based the calculated distance. These features $f_{sm}$ are treated as similar feature sets $\mathbb{R}_{sm}$. 
  At  the  last,  we 
     adaptability select  the  most  appropriate
      benchmark $f_b$  from  $\mathbb{R}_{sm}$ by  the  overall  (mean),  maximum  (maximum) 
       and most  moderate (median)  aspects  and  use  this  benchmark
         to refine the extracted spatio-temporal features $f_{st}$.
     Fig. \ref{fig1} shows the complete feature refinement process. We can clearly see that after
      refinement by GDA,
      the intra-class similarity and inter-class dissimilarity of $f_{st}$
     have greatly enhanced.
     Next, we will introduce GDA in detail from the following five parts:
      Unlabeled gait data,  Adjustment set, Similar feature set, Benchmark and Refinement.

\subsubsection{Unlabeled gait data}
 The selection of unlabeled gait data $V_{useqs}$ directly determines the robustness
  of the adjustment set $\mathbb{R}_S$.
 It should be ensured that the unlabeled gait data $V_{useqs}$ should
  contain the appropriate number of subjects.
  If the number is small, it will not be representative enough, and if the number is large,
   it will lead to problems similar to overfitting. Fig. \ref{fig4} of the experimental section verifies
  this conclusion. Another thing to note is that subjects in unlabeled gait data $V_{useqs}$ should have an appropriate
  sex ratio and a reasonable age distribution.

\subsubsection{Adjustment set}
  When unlabel gait data $V_{useqs}$ is found, we will use the trained SFE model
  to extract their spatio-temporal gait features and use these features as adjustment set $\mathbb{R}_S$.
   The formula 
   for summarizing $\mathbb{R}_S$ is defined as:
  \begin{equation}
    \mathbb{R}_S = SFE(V_{useqs}),
    \end{equation}
    where $V_{useqs}$ are the gait sequences from the unlabeled gait data, which contains the appropriate
    number of
    subjects. 
    $\mathbb{R}_S$ is the adjustment set used to refine the spatio-temporal gait features $f_{st}$.

\subsubsection{Similar feature set}
 When we get the adjustment set $\mathbb{R}_S$ from unlabeled gait data $V_{useqs}$,
 we first calculate the distance between these features in $\mathbb{R}_S$ and extracted gait spatio-temporal
   features $f_{st}$, and  then select  some  features $f_{sm}$ similar  to
   $f_{st}$ from $\mathbb{R}_S$ by the calculated distance. These features $f_{sm}$ are treated as
    similar feature set $\mathbb{R}_{sm}$.
  The formula for selecting the similar features $f_{sm}$ is defined as:
\begin{equation}
  \begin{aligned}
  P\{d(f_{sm},&f_{st})- d(f_{rs},f_{st}))<0\}=10^{-t}, \\
  f_{rs} \in &\mathbb{R}_S, f_{sm} \in \mathbb{R}_{sm}, \mathbb{R}_{sm} \in  \mathbb{R}_S,
  \end{aligned}
  \end{equation}
  where $f_{st}$ is the spatio-temporal feature extracted from raw gait sequences $V_{seqs}$ in the
   probe set or gallery set, $f_{sm}$ is the spatio-temporal feature in $\mathbb{R}_{sm}$.
   $f_{rs}$ is the spatio-temporal feature in $\mathbb{R}_S$.  $d(f_{sm},f_{st})$ represents the distance
    between $f_{sm}$
    and $f_{st}$.
   $d(f_{rs},f_{st})$ represents the distance between $f_{rs}$ and $f_{st}$.
   The distances mentioned here are measured by the Euclidean norm.
   This formula shows that the probability that the distance between $f_{sm}$ and $f_{st}$ is smaller than the distance
   between $f_{rs}$ and $f_{st}$ is $10^{-t}$. It can help us find some features $f_{sm}$, which
    most similar to extracted spatio-temporal features $f_{st}$ from $\mathbb{R}_S$. 

    \subsubsection{Benchmark}   
   when we get $\mathbb{R}_{sm}$, we will  
   select  the  most  appropriate  benchmark $f_b$ to achieve the best refinement effect.
      The formula for determining $f_b$ is defined as:
        \begin{equation}
          \begin{aligned}
          global = (max(\cdot) &+ mean(\cdot) + median(\cdot)) / 3,\\
          f_b &= global(f_{sm}), \\ 
          &f_{sm} \in \mathbb{R}_{sm},
          \end{aligned}
          \end{equation}
    where $\mathbb{R}_{sm}$ has some features similar to the extracted saptio-temporal gait
     feature $f_{st}$. 
     $max$, $mean$, $median$ respectively represent the maximum, average, and median values. $global(\cdot)$ can
     ensure that all similar features in $\mathbb{R}_{sm}$ are fully utilized, thus ensuring robustness of the
     benchmark
     $f_b$. It is worth noting that for each extracted gait feature $f_{st}$,
     the corresponding $f_b$ is different.

     \subsubsection{Refinement}   
    When the appropriate benchmark $f_b$ is calculated,
    we can use it to refine the previously extracted spatio-temporal
    gait  features $f_{st}$.
      This operation can make these features have hign intra-class similarity and inter-class dissimilarity,
      and make them more discriminative.
      The feature $f_{st}$ extracted in the test stage is divided into gallery set and
       probe set.
      When only the features $f_g$ in probe set are refined, the specific formula is defined as follows:
     \begin{equation}
      \begin{aligned}
      d'(f_g,f_q) = d(&f_g,f_q) - \lambda_q\cdot d(f_q,f_b),\\
      f_g \cup& f_q = f_{st},
      \end{aligned}
      \end{equation}
      where $d(f_g,f_q)$ is the original distance between features $f_q$ in the probe set and
      features $f_g$ in the gallery set, it is an important reference to determine the identity of subject.
       $d(f_q, f_b)$ is the value that spatio-temporal feature $f_q$  need to be refined,
        $d'(f_g,f_q)$ is the refined distance between $f_q$ in the probe set and
        $f_g$ in the gallery set.  However,
       the formula only refines the features in probe set to modify the final distance for feature matching.
        When we want to
        refine the spatio-temporal gait features 
        in both the gallery set and probe set,
        the formula can be defined as follows:
        \begin{equation}
          d'(f_g,f_q) = d(f_g,f_q) - \lambda_g\cdot d(f_g, f_b),-\lambda_q\cdot d(f_q, f_b),
          \end{equation}
          where  $d(f_g,f_b)$ is the value used to refine the spatio-temporal features $f_g$
           in gallery set.
            $\lambda_g$ and $\lambda_q$ are hyperparameters that determine the degree of refinement.
           Extensive experiments have proved that GDA has the best effect when both $\lambda_g$ and $\lambda_q$ are set to 0.5.

                \begin{table*}[]
                  \centering
                  \renewcommand\arraystretch{1.3}
                  \setlength{\tabcolsep}{3.0mm}
                  \caption{Averaged rank-1 accuracies on CASIA-B,excluding identical-view cases.}
                  \label{tab2}
                  \begin{tabular}{|l|c|c|c|c|c|c|c|c|c|c|c|c|c|}
                  \hline
                  \multicolumn{2}{|c|}{Gallery NM\#1-4}                    & \multicolumn{11}{c|}{0°-180°}                                                                                                                                                                                                                                                                                     & \multirow{2}{*}{mean} \\ \cline{1-13}
                  \multicolumn{2}{|c|}{Probe}                              & \multicolumn{1}{l|}{0°}   & \multicolumn{1}{l|}{18°}  & \multicolumn{1}{l|}{36°}  & \multicolumn{1}{l|}{54°}  & \multicolumn{1}{l|}{72°}  & \multicolumn{1}{l|}{90°}  & \multicolumn{1}{l|}{108°} & \multicolumn{1}{l|}{126°} & \multicolumn{1}{l|}{144°} & \multicolumn{1}{l|}{162°} & \multicolumn{1}{l|}{180°} &                       \\ \hline
                  \multicolumn{1}{|c|}{\multirow{8}{*}{NM}} & CNN-LB \cite{wu2017a}         & 82.6                      & 90.3                      & 96.1                      & 94.3                      & 90.1                      & 87.4                      & 89.9                      & 94.0                      & 94.7                      & 91.3                      & 78.5                      & 89.9                  \\ 
                  \multicolumn{1}{|c|}{}                    & GaitSet1 \cite{chao2019gaitset}    & 90.8                      & 97.9                     & 99.4             & 96.9                      & 93.6                      & 91.7                      & 95.0             & 97.8                      & 98.9                      & 96.8             & 85.8                      & 95.0                  \\ 
                  \multicolumn{1}{|c|}{}                    & Gait-Joint \cite{song2019gaitnet}     & 75.6                      & 91.3                   & 91.2             & 92.9                      & 92.5                      & 91.0                      & 91.8             & 93.8                      & 92.9                      & 94.1             & 81.9                      & 89.9                  \\ 
                  \multicolumn{1}{|c|}{}                    & GaitNet \cite{zhang2019gait}     & 91.2                      & 92.0                      & 90.5                      & 95.6                      & 86.9                      & 92.6                     & 93.5                      & 96.0                      & 90.9                      & 88.8                      & 89.0                      & 91.6                  \\  
                  \multicolumn{1}{|c|}{}                    & CNN-Ensemble \cite{wu2017a} & 88.7                      & 95.1                      & 98.2                      & 96.4                      & 94.1                      & 91.5                      & 93.9                      & 97.5                      & 98.4                      & 95.8                      & 85.6                      & 94.1                  \\  
                  \multicolumn{1}{|c|}{}                    & GaitSet2 \cite{chao2021gaitset} & 91.1                      & 99.0                      & \textbf{99.9}                      & 97.8                      & 95.1                      & 94.5                      & 96.1                      & 98.3                      & 99.2                      & 98.1                      & 88.0                      & 96.1                  \\  
                  \multicolumn{1}{|c|}{}                    & GaitPart \cite{Fan_2020_CVPR}     & 94.1                      & 98.6                      & 99.3                      & \textbf{98.5}                      & 94.0                      & 92.3                        & 95.9                      & 98.4                      & 99.2                        & 97.8                      & 90.4                      & 96.2                 \\  \cline{2-14}
                  \multicolumn{1}{|c|}{}                    & Ours    & \textbf{95.3}                         & \textbf{99.2}                      & 99.1                      & 98.3                     & \textbf{95.4}                      & \textbf{94.4}                        & \textbf{96.5}                      & \textbf{98.9}                      & \textbf{99.4}                      & \textbf{98.2}                     & \textbf{92.0}                      & \textbf{97.0}                  \\ \hline
                  \multirow{7}{*}{BG}                       & CNN-LB  \cite{wu2017a}      & 64.2                      & 80.6                      & 82.7                      & 76.9                      & 64.8                      & 63.1                      & 68.0                      & 76.9                      & 82.2                      & 75.4                      & 61.3                      & 72.4                  \\  
                                                            & GaitSet1  \cite{chao2019gaitset}     & 83.8                      & 91.2                      & 91.8                      & 88.8                      & 83.3                      & 81                      & 84.1                      & 90.0                      & 92.2                      & 94.4             & 79.0                      & 87.2                  \\ 
                                                            & GaitSet2  \cite{chao2021gaitset}     & 86.7                      & 94.2                      & 95.7                      & 93.4                      & 88.9                      & 85.5                      & 89.0                      & 91.7                      & 94.5                      & \textbf{95.9}             & 83.3                      & 90.8                  \\
                                                            & MGAN   \cite{he2019multi}        & 48.5                      & 58.5                      & 59.7                      & 58.0                      & 53.7                      & 49.8                      & 54.0                      & 61.3                      & 59.5                      & 55.9             & 43.1                      & 54.7                  \\ 
                                                            & GaitNet \cite{zhang2019gait}      & 83.0                      & 87.8                      & 88.3                      & 93.3             & 82.6                      & 74.8                      & 89.5             & 91.0             & 86.1                      & 81.2                      & 85.6             & 85.7                  \\  
                                                            & GaitPart \cite{Fan_2020_CVPR}      & 89.1                      & 94.8                      & 96.7                      & \textbf{95.1}             & 88.3                      & \textbf{94.9}                      & 89.0             & 93.5             & 96.1                      & 93.8                      & 85.8             & 91.5                  \\ \cline{2-14}                                           
                                                            & Ours    & \textbf{91.3}                     & \textbf{94.9}                      & \textbf{95.5}                      & 93.64                      & \textbf{90.5}                      & 84.4                      & \textbf{90.8}                      & \textbf{95.8}                       & \textbf{97.6}                      & 94.14                      & \textbf{88.0}                      & \textbf{92.4}                  \\ \hline 
                  \multirow{6}{*}{CL}                       & CNN-LB  \cite{wu2017a}      & 37.7                      & 57.2                      & 66.6                      & 61.1                      & 55.2                      & 54.6                      & 55.2                      & 59.1                      & 58.9                      & 48.8                      & 39.4                      & 54.0                  \\   
                                                            & GaitSet1  \cite{chao2019gaitset}    & 61.4                      & 75.4                      & 80.7                      & 77.3                      & 72.1                      & 70.1                      & 71.5                      & 73.5                      & 73.5                      & 68.4                     & 50.0                      & 70.4                  \\ 
                                                            & GaitSet2  \cite{chao2021gaitset}    & 59.5                      & 75.0                      & 78.3                      & 74.6                      & 71.4                      & 71.3                      & 70.8                      & 74.1                      & 74.6                      & 69.4                     & 54.1                      & 70.3                  \\    
                                                            & MGAN   \cite{he2019multi}   & 23.1                      & 34.5                      & 36.3                      & 33.3                      & 32.9                      & 32.7                      & 34.2                      & 37.6                      & 33.7                      & 26.7             & 21.0                      & 31.5                  \\ 
                                                            & GaitNet \cite{zhang2019gait}        & 42.1                      & 58.2                      & 65.1                      & 70.7                      & 68.0                      & 70.6             & 65.3                      & 69.4                      & 51.5                      & 50.1                      & 36.6                      & 58.9                  \\ \cline{2-14} 
                                                            & Ours    & \multicolumn{1}{c|}{\textbf{73.0}} & \multicolumn{1}{c|}{\textbf{86.4}} & \multicolumn{1}{c|}{\textbf{85.6}} & \multicolumn{1}{c|}{\textbf{82.7}} & \multicolumn{1}{c|}{\textbf{76.8}} & \multicolumn{1}{c|}{\textbf{74.3}} & \multicolumn{1}{c|}{\textbf{77.1}} & \multicolumn{1}{c|}{\textbf{80.7}} & \multicolumn{1}{c|}{\textbf{79.6}} & \multicolumn{1}{c|}{\textbf{77.6}} & \textbf{64.7}             & \textbf{78.0}                  \\ \hline
                  \end{tabular}
                  \end{table*}

                  \begin{table}[h]
                      
                    \centering
                    \renewcommand\arraystretch{1.3}
                    \setlength{\tabcolsep}{3.0mm}
                    \caption{Averaged rank-1 accuracies on mini-OUMVLP, excluding identical-view cases.}
                    \label{tab3}
                    \begin{tabular}{|c|c|c|}
                    \hline
                    \multirow{2}{*}{probe} & \multicolumn{2}{c|}{Gallery All 14 Views} \\ \cline{2-3} 
                                           & GaitPart(experiment)          & Ours                   \\ \hline
                    0°                     & 73.5             & \textbf{77.0}          \\ \hline
                    15°                    & 79.4             & \textbf{84.0}          \\ \hline
                    30°                    & 87.2             & \textbf{88.5}          \\ \hline
                    45°                    & 87.1             & \textbf{88.4}          \\ \hline
                    60°                    & 83.0             & \textbf{85.6}          \\ \hline
                    75°                    & 84.2             & \textbf{87.3}          \\ \hline
                    90°                    & 81.5             & \textbf{85.7}          \\ \hline
                    180°                   & 73.6             & \textbf{78.4}          \\ \hline
                    195°                   & 79.8             & \textbf{84.7}          \\ \hline
                    210°                   & 87.3             & \textbf{88.0}          \\ \hline
                    225°                   & 88.3             & \textbf{89.3}          \\ \hline
                    240°                   & 83.8             & \textbf{86.2}          \\ \hline
                    255°                   & 84.1             & \textbf{86.1}          \\ \hline
                    270°                   & 82.0             & \textbf{86.3}          \\ \hline
                    mean                   & 82.5             & \textbf{85.4}          \\ \hline
                    \end{tabular}
                    \end{table}

  \section{Experiments}

  There are four parts to our experiment: the first part is an introduction of
  CASIA-B, mini-OUMVLP and training details. The second part shows
  the results of our method compared with other state-of-the-art methods. The third part is the ablation
  experiment, which describes the effectiveness of each module of the proposed method.
    The last part is an additional experiment,
    which presents the application of the proposed method on the cross dataset.
  \subsection{Datasets and Training Details}
 
 {\bf CASIA-B} \cite{zheng2011robust} is the main database to test the effectiveness of the gait method.
 It has a total of $124 \times 10 \times 11=13640$ sequences,
  which 124 mean the number of subjects, 10 mean the groups of each subject, 11 mean angles of each group.
  Its walking conditions include walking in coats (CL), walking with a bag (BG) and  normal walking (NM).
  CL has 2 groups, BG has 2 groups,
 NM has 6 groups. 11 angles include 0°, 18°, 36° 54°, 72°, 90°, 108°, 126°, 144°, 162°, 180°.
 The first
 74 subjects of CASIA-B are used as the training set and 
 the last 50 subjects are considered as the test set. The test set is divided into gallery set and probe set.
 The first 4 sequences 
  in NM condition  is set to gallery set, and the rest sequences are divided into 3
 probe sets: BG \#1-2,  CL \#1-2, NM \#5-6.
 
 {\bf mini-OUMVLP} \cite{chen2021effective} evolved from OU-MVLP \cite{takemura2018multi}.
 The number of original OU-MVLP is too 
 large and the requirement
  of GPU is high, so we choose the first 500 subjects
  of OU-MVLP 
 as our dataset: mini-OUMVLP. Each subject of mini-OUMVLP has 2 groups (\#00-01), each group has
  14 views (0°, 15°, ..., 90°; 180°, 195°, ..., 270°). The first 350 subjects of mini-OUMVLP are used as
  the training set, the rest 150 subjects are used as the test set.
  Seq \#01 in the test set is set to gallery set and Seq \#00 in the test set is set to probe set.

 In all experiments, we crop the size of input silhouettes to 64 × 44 and consider 50 gait silhouettes as
  a gait sequence.
  We use adam as the optimizer for the model. 
  The margin of hard triplet loss is set to 0.2. 
 
  {\bf1)} In the dataset CASIA-B, we train the proposed method 78K iterations. GaitSet has been trained
   on 80K iterations. 
   GaitPart has been trained on 120k iterations.
  The parameter settings of GaitPart and GaitSet are followed as their original papers.
   The parameters $p$ and $k$ are set to 8 and 16 respectively. The
 number of channels in both $C$1 and $C$2 is set as 32, $C$3 is set as 64, $C$4 is set as 128.
  The block for FFE is set as 4. The learning rate of this module is set as 1$e$-4.
  We simulate real-life unlabeled gait data with gait sequences from the test set and extract
  the adjustment set
  by the proposed Spatio-temporal Feature Extraction module.
  The probability for selecting similar features is set to 0.001.
 
  {\bf2)} In the dataset mini-OUMVLP, we train the proposed method 120K iterations.
   GaitPart has been trained on 120k iterations.
   The parameter settings of GaitPart are followed as its original paper. The number of channels  
  in both $C$1 and $C$2 is set as 32, $C$3 is set as 64, $C4$  is set as 128.
   The number of blocks for FFE is set as 4. 
   The parameters $p$ and $k$ are set to 8 and 16 respectively.
    The learning rate of this module is also set as 1$e$-4. 
  We also simulate real-life unlabeled gait data with gait sequences from the test set and extract
  the adjustment set
  by the proposed Spatio-temporal Feature Extraction module.
  The probability for selecting similar features  is set to 0.001.

  {\bf3)} In additional experiments, we train proposed method 120k iterations on CASIA-B. The batch size
   is set to 8 $\ast$ 16. The
  numbers of channels for both $C$1 and $C$2 is set as 32, $C$3 is set as 64, $C$4 is set as 128.
   The block for FFE is set as 4. The learning rate of this module is also set as 1$e$-4. We test
   the proposed method on mini-OUMVLP.
   The first 350 subjects in mini-OUMVLP are simulated as unlabeled gait data in real life, and the next
   150 subjects are used as the test set. The probability for selecting similar features  is set to 0.001.

   \begin{table*}[]
    \centering
    \renewcommand\arraystretch{1.3}
    \setlength{\tabcolsep}{2.5mm}
    \caption{Averaged rank-1 accuracies on CASIA-B,excluding identical-view cases.}
    \label{tab4}
    \begin{tabular}{|l|c|c|c|c|c|c|c|c|c|c|c|c|c|}
    \hline
    \multirow{2}{*}{probe}               &\multicolumn{6}{|c|}{Fine-Grained Feature Extraction (FFE)}  &\multicolumn{3}{|c|}{Feature alignment} &\multicolumn{3}{|c|}{walking conditions}\\ \cline{2-13}
                                 & \multicolumn{1}{l|}{8blocks}   & \multicolumn{1}{l|}{4blocks}  & \multicolumn{1}{l|}{2blocks}  & \multicolumn{1}{l|}{1blocks}  & \multicolumn{1}{l|}{4blocks+8blocks}  & \multicolumn{1}{l|}{4blocks+4blocks}  & \multicolumn{1}{l|}{0.001} & \multicolumn{1}{l|}{0.01} & \multicolumn{1}{l|}{0.1} & \multicolumn{1}{l|}{NM} & \multicolumn{1}{l|}{BG} &  \multicolumn{1}{l|}{CL}                     \\ \hline
    \multicolumn{1}{|c|}{a}         & \checkmark                     &                      &                      &                     &                       &                       &                       &                       &                       & 94.5                     &   86.8                    &  75.4                 \\ \hline
    \multicolumn{1}{|c|}{b}                          &                      & \checkmark                     &             &                     &                      &                       &              &                      &                       & 95.5            &90.1                     &  74.4               \\             \hline
    \multicolumn{1}{|c|}{c}                         &                      &                    &  \checkmark            &                      &                       &                       &              &                       &                       &  95           & 87.4                      &  70.6                 \\ \hline
    \multicolumn{1}{|c|}{d}                          &                       &                       &                      &   \checkmark                    &                       &                       &                       &                       &                       & 95                      & 86.4                      & 70.2                 \\  \hline
    \multicolumn{1}{|c|}{e}                      &                      &                       &                      &                      &  \checkmark                      &                       &                       &                      &                      &   94.7                   &  85.6                     &  71.9                 \\  \hline
    \multicolumn{1}{|c|}{f}                          &                      &                      &                      &                     &                       & \checkmark                         &                      &                       &                        &   95.0                  & 84.6                     &  64.4               \\  \hline
    \multicolumn{1}{|c|}{g}                        &                     & \checkmark                    &                      &                      &                       &                         &     \checkmark                  &                       &                       &   \textbf{97.0}                     &  \textbf{92.4}                    & \textbf{78.0}                   \\ \hline
    \multicolumn{1}{|c|}{h}                       &                      & \checkmark                     &                      &                      &                      &                      &                       &  \checkmark                     &                       &   96.1                   & 91.5                   &  76.3                \\  \hline
    \multicolumn{1}{|c|}{i}                                                &                      & \checkmark                     &                       &                      &                       &                       &                       &                       &   \checkmark                    & 96             &  90.8                    & 75.6                 \\ \hline
                                                   
    \end{tabular}
    \end{table*}

\subsection{Results}

  We have done a lot of experiments on CASIA-B and mini-OUMVLP, and the results proved that the
  proposed method has better performance than some state-of-the-art methods.

{\bf CASIA-B}. Tab. \ref{tab2} shows the results of some experiments with our method and some state-of-the-art
 methods. 
 Except for our method, we also use GaitSet and GaitPart to do experiments on CASIA-B and the results of other methods are
 directly taken from their original papers. 
 On normal walking conditions (NM), our accuracy at all views is greater than 90$\%$, and the 
 average accuracy can achieve 97.0$\%$, even better than Gaitpart. On walking with a bag (BG), the average
 accuracy can achieve 92.4$\%$, and only two views (90°, 180°) are below 90$\%$. On walking in coats
 (CL), the average accuracy can achieve 78.0$\%$, and we are the best on all views
  compared with other state-of-the-art works.

{\bf mini-OUMVLP}. In order to verify the robustness of our method, 
 we try to do some experiments on another larger data set: mini-OUMVLP. 
 Tab. \ref{tab3} shows the results of our experiments. Because the original OU-MVLP dataset has so many
 subjects that it requires a lot of CPU resources,
 we use a smaller version that contained only the first 500 subjects of OU-MVLP. 
 We keep the parameter settings of GaitPart as their original experiment. It
 should be notice that the original data of mini-OUMVLP are incomplete 
 , so the identification accuracy will be a little affected. 
  The average
 accuracy of all views of our method
 is 85.4$\%$, even better than GaitPart, which accuracy is 82.5$\%$. We are proud to have a
  higher performance than GaitPart
 in all 14 views.

 \begin{table}[h]
            
  \centering
  \renewcommand\arraystretch{1.3}
  \setlength{\tabcolsep}{2.5mm}
  \caption{Averaged rank-1 accuracies on mini-OUMVLP, excluding identical-view cases.}
  \label{tab5}
  \begin{tabular}{|c|c|c|c|}
  \hline
  \multirow{2}{*}{probe} & \multicolumn{2}{c|}{Gallery All 14 Views} \\ \cline{2-3} 
                                & Ours(Adjustment)       & Ours(No Adjustment)               \\ \hline
  0°                            & \textbf{74.3}         & 70.8     \\ \hline
  15°                           & \textbf{81.6}         & 79.9     \\ \hline
  30°                           & 87.1                  & \textbf{87.3}     \\ \hline
  45°                           & 87.0                  & \textbf{87.1}     \\ \hline
  60°                           & \textbf{83.2}         & 82.5     \\ \hline
  75°                           & \textbf{85.0}         & 83.8     \\ \hline
  90°                           & \textbf{83.0}           & 82.3     \\ \hline
  180°                          & \textbf{76.4}         & 73.6     \\ \hline
  195°                          & \textbf{81.8}         & 80.6     \\ \hline
  210°                          & \textbf{86.1}                  & 85.9     \\ \hline
  225°                          & 86.2                  & \textbf{86.8}    \\ \hline
  240°                          & \textbf{84.0}           & 83.2     \\ \hline
  255°                          & \textbf{83.0}                    & 82.4     \\ \hline
  270°                          & \textbf{83.0}           & 81.8     \\ \hline
  mean                          & \textbf{83.0}           & 82.0     \\ \hline
  \end{tabular}
  \end{table}

 \subsection{Ablation Experiments}

 To prove the role of every part of the proposed method. We perform several ablation experiments
  on CASIA-B with different parameters, including setting different numbers of the blocks in Fine-grained Feature 
  Extraction (FFE) and
  the different probability in Global Distance Alignment (GDA). The results are shown as Tab. \ref{tab4}. We will
  analyze these results in the next part. 

{\bf Effectiveness of FFE}. In order to explore how fine-grained features can be adequately extracted. 
 We do different experiments on the number of blocks of FFE. 8 blocks, 4 blocks,
 2 blocks, and 1 block mean that we divide the most representative temporal feature into 8 parts, 4 parts, 2 parts, and 1 part to extract the fine-grained
 spatial feature. 4 blocks + 8 blocks means that we divide the most representative temporal feature into
  4 parts for
  fine-grained feature extractions,
  and then divide it into 8 parts again for further extraction of the fine-grained feature.
   4 blocks + 4 blocks means that we divide the most representative temporal feature
  into 4 parts to learn the fine-grained features and then divide it into 4 parts again to further
  extract fine-grained features. {\bf1)} By comparing the groups a, b, c, d, we find that  when we divide
  the most representative temporal features into 4 parts to extract the fine-grained features, it shows the best performance.
   This can prove that when we use the partial idea to extract fine-grained features, it is necessary to block the feature map to 
   appropriate parts. If
   the most representative temporal feature is divided too fine, the information will be repeatedly extracted,
    and if the feature graph is
    divided too coarse, the fine-grained information will not be fully extracted.
   Both of them will affect the accuracy of recognition. {\bf2)} By comparing groups b, e, and f, we can observe that when we 
   divide the most representative temporal feature into 4 parts and extract fine-grained features only once can
   get the best
    result. This can prove that the most representative temporal features extracted by Temporal Feature Fusion (TFF) are high-level features,
     so it only needs to block once and then extract fine-grained features once can fully obtain fine-grained features,
      and too much extraction will lead to information loss,
    which has a great impact on recognition accuracy. The results of the experiments can also show the advantages of our Spatio-temporal Feature
     Extraction module: spatial features can be fully extracted by using only one blocking operation
     in high-level features, which simplifies the structure of the network.

  {\bf Effectiveness of GDA}.
   GDA is not to optimize the network structure, it belongs to the post-processing of data,
   it uses  unlabeled data set as the benchmark
   to refine the spatio-temporal features in the probe set and gallery set to
    make them have low inter-class similarity and high
    intra-class similarity.
   We will prove its 
   effectiveness through some experiments.
    The most important part of GDA is to select the appropriate benchmark from the adjustment set $\mathbb{R}_S$
     to
    refine spatio-temporal features $f_{st}$ in gallery set and probe set. The most important factor
     to select an appropriate benchmark 
    is to find some features $\mathbb{R}_{sm}$ similar to spatio-temporal features $f_{st}$ from $\mathbb{R}_S$.
     We will experiment to find out how many similar features we should find to have the best performance.
  0.1, 0.01, 0.001 are the important similarity indicator to find similar feature set $\mathbb{R}_{sm}$.
  These mean that the probability that the distance between 
  $f_{sm}$ in $\mathbb{R}_{sm}$ and $f_{st}$ is smaller than the distance between 
   $f_s$ in $\mathbb{R}_S$ and $f_{st}$ are 0.1, 0.01, 0.001 respectively. {\bf1)}
  By comparing 
  the groups g, h, and i, it can be clearly observed that  when we set the probability to  0.001,
   the result will be best.
   This can prove that only a small part of the data in $\mathbb{R}_S$ is related to the
    spatio-temporal features $f_{st}$,
   So we can not set the possibility too high. {\bf2)}
   By comparing the groups
    b and g, we can 
   see the significance of GDA, the accuracies on NM, BG, and CL both get a good promotion after being refined by 
   GDA. 
   This fully proves that when we use GDA to refine the spatio-temporal features $f_{st}$
   can make them have low inter-class similarity and high intra-class similarity, thus increasing their discrimination.

   \subsection{Additional Experiments}
   
   The training phase and test phase are all under the same data set on previous experiments,
    and the unlabeled data used for GDA is the test set. To further 
   prove the robustness of the GDA module, we try to experiment on cross dataset.
    We first train the proposed Spatio-temporal Feature Extraction (SFE) module on CAISA-B, when the network parameters are trained,
    we do the test on mini-OUMVLP.
    We use the gait sequences of the first 350 subjects of mini-OUMVLP as the unlabeled gait sequences $V_{useqs}$, and then use the
    trained SFM module to extract their spatio-temporal gait features as the adjustment set $\mathbb{R}_S$.
    Finally, we use the rest 150 subjects as the test
    set to test the effectiveness of GDA.
    Tab. \ref{tab5}  presents our results on mini-OUMVLP.
    Adjustment means that we refine the extracted spatio-temporal gait features  in the
     probe set and gallery
     set by GDA before the test, and No means test without the refinement by GDA. It can 
    be seen that after the processing of GDA, the recognition accuracies can be effectively improved.
     This cross dataset experiment can further
    prove the robustness of GDA.

  \begin{figure}[htbp]
      \centering
      \includegraphics[width=9cm]{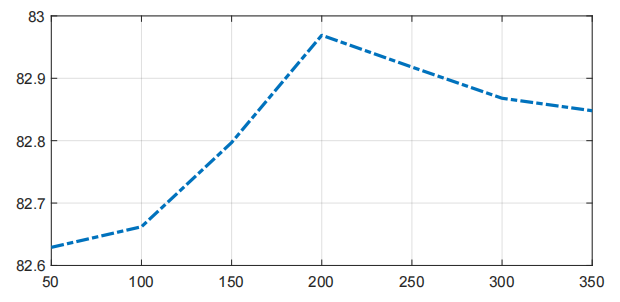}
      \caption{The relationship between the number of subjects used to global distance alignment
       and the final recognition accuracy.}
      \label{fig4}
  \end{figure} 

  We also do some experiments to explore how many subjects should be used as the unlabeled data set.
  Fig. \ref{fig4} shows how the final recognition accuracy varies with the number of subjects
   who are used to as the unlabeled gait data. We can observe that at the beginning,
 with the increase of the number of subjects, the accuracy of recognition becomes higher and higher.
 But as the number of subjects reaches a threshold,
    if we continue to increase the number of subjects, the recognition accuracy will decrease.
    This shows that we should choose the appropriate number of subjects when we find the unlabeled gait data,
     too few subjects may lead to the unlabeled data lacking representativeness,
      and too many subjects may lead to problems like overfitting.

\section{Conclusion}
In this paper, we propose an interesting idea of using global distance alignment techniques to
 enhance the discrimination of features.
  Thus our method is proposed. It first uses Fine-grained Feature Extraction (FFE) and
 Temporal Feature Fusion (TFF) to effectively learn the spatio-temporal features
 from raw 
 gait silhouettes, then uses a large  number of unlabeled gait data in
 real life as a benchmark to refine the extracted spatio-temporal
 features to make them have low inter-class similarity and high
 intra-class similarity, thus enhancing their discrimination.
  Extensive experiments have proved that our method
 has a great performance on two main gait datasets: mini-OUMVLP and CASIA-B.
  In the future, we will try to introduce hyperspectral data sets and radar data sets to extract
  robust features for the task of gait recognition.


%

\ifCLASSOPTIONcaptionsoff
  \newpage
\fi

\bibliographystyle{IEEEtran}
\bibliography{refs}
\end{document}